\documentclass[10pt,twocolumn,letterpaper]{article}

\usepackage{cvpr}
\usepackage{times}
\usepackage{epsfig}
\usepackage{graphicx}
\usepackage{amsmath}
\usepackage{amssymb}
\usepackage{subfigure}
\usepackage{color}
\usepackage{setspace}
\usepackage{algorithm}
\usepackage{algorithmic}
\usepackage{multirow}
\usepackage[L7x,T1]{fontenc}
\usepackage[utf8]{inputenc}
\usepackage[lithuanian,english]{babel}
\DeclareMathOperator*{\argmax}{argmax}

\usepackage[pagebackref=true,breaklinks=true,letterpaper=true,colorlinks,bookmarks=false]{hyperref}

\cvprfinalcopy 

\def\cvprPaperID{1736} 

\ifcvprfinal\fi

\begin{document}
\title{Combining Word Embeddings and N-grams for Unsupervised Document Summarization}

\author{Zhuolin Jiang, Manaj Srivastava, Sanjay Krishna Gouda, David Akodes, Richard Schwartz \\
 Raytheon BBN Technologies, Cambridge,  MA, 02138 \\
 {\tt\small  \{zhuolin.jiang, manaj.srivastava, sanjaykrishna.gouda, david.akodes, rich.schwartz\}@raytheon.com}
}

\maketitle

\begin{abstract}
Graph-based extractive document summarization relies on the quality of the sentence similarity graph. Bag-of-words or tf-idf based sentence similarity uses exact word matching, but fails to measure the semantic similarity between individual words or to consider the semantic structure of sentences. In order to improve the similarity measure between sentences, we employ off-the-shelf deep embedding features and tf-idf features, and introduce a new text similarity metric. An improved sentence similarity graph is built and used in a submodular objective function for extractive summarization, which consists of a weighted coverage term and a diversity term. A Transformer based compression model is developed for sentence compression to aid in document summarization. Our summarization approach is extractive and unsupervised. Experiments demonstrate that our approach can outperform the tf-idf based approach and achieve state-of-the-art performance on the DUC04 dataset, and comparable performance to the fully supervised learning methods on the CNN/DM and NYT datasets.
\end{abstract}

\section{Introduction}

State-of-the-art summarization performance has been achieved by using supervised learning methods, which are mainly based on neural network architectures and require a large corpus of document-summary pairs ~\cite{nallapati2016abstractive},~\cite{see2017get},~\cite{narayan2018ranking} and~\cite{gehrmann2018bottom}.
Alternative approaches to document summarization employ unsupervised techniques~\cite{Takamura:2009,mihalcea2004textrank,erkan2004lexrank,Lin:2011,Lin:2010}.
Those include graph-based extractive summarization methods, such as~\cite{Lin:2011,erkan2004lexrank,mihalcea2004textrank}, which require a similarity graph between sentences as input to the summarization system.
The similarity between sentences is usually computed using bag-of-words or tf-idf features, which do not incorporate similarity in the semantics.
Modeling sentence semantic similarity is challenging because of the variability of linguistic expression, where different words in different orders can express the same meanings, or the same set of words in different orders can express totally different meanings. Due to this, traditional sparse and hand-crafted features such as bag-of-words and tf-idf vectors fail to effectively capture the similarity between individual words and semantic structure (and context) of sentences.
Alternatively, distributed semantic representations (or word embeddings) of each word, such as word2vec~\cite{Mikolov:2013} and GloVe~\cite{pennington-etal-2014-glove}, do a better job of capturing the word or sentence level semantics, and have been widely used in many NLP tasks.~\cite{kaageback2014extractive} and~\cite{rossiello:2017} represent the embedding of a sentence by averaging the embedding vectors for each word in the sentence. But there is limited work that uses these deep word embeddings in an unsupervised setting for extractive document summarization.~\cite{kobayashi:2015} introduces a summarization method that estimates KL-divergence between the document and its summary based on embedding distributions.

In this paper, we explore two popular deep embedding features: word2vec and BERT~\cite{Devlin:2018} in a submodular framework for document extractive summarization. Our document summarization framework is unsupervised and is therefore useful for the case of limited or no document-reference summary pairs. In order to use the strengths of these two types of features, we combine them to further improve the similarity measure. In addition, we investigate the effect of using abstractive sentence compression for extractive document summarization. Towards that end, we train a Transformer model~\cite{Vaswani:2017} to compress the sentences from a document before performing submodular sentence selection. Our main contributions are:
\begin{itemize}
\item We improve the sentence similarity graph by exploiting off-the-shelf neural word embedding models for graph-based submodular sentence selection, where a similarity graph for pair-wise sentences is required. We provide thorough experimental comparisons between different sentence similarity measures.
\item We show that combining off-the-shelf neural word embeddings and tf-idf features can improve the performance of document summarization.
\item We show that a Transformer based sentence compression method can improve the performance of document summarization.
\end{itemize}

\section{Unsupervised Document Summarization}

Similar to~\cite{Lin:2011}, we extract a subset of sentences $A$ from the whole set of sentences $V$ in a document $D$ as the summary by maximizing a submodular objective function.

\subsection{Similarity Graph Construction}

Given a document $D$, we construct an undirected similarity graph $G=(V, E)$, where the vertices $v\in V$ are sentences in $D$ and the edges $e\in E$ model pairwise relation between the sentences. The weight $w_{i,j}$ associated with the edge $e_{i,j}$ measures the similarity between vertices or sentences $v_i$ and $v_j$. $w_{i,j}$ is computed as: $w_{i,j}=exp(-\beta d^2(x_i, x_j))$, where $x_i$ is the feature descriptor of $v_i$ and $d(x_i, x_j)$ measures the difference between $x_i$ and $x_j$. As suggested in~\cite{Zelnik-Manor:2004}, we set the normalization factor to $\beta=1/\delta_i\delta_j$, and select the scaling parameter $\delta_i$ for $v_i$ through the local statistic of $v_i$'s neighborhood. We set $\delta_i=d(x_i,x_k)$ where $x_k$ corresponds to the $k$-th nearest neighbor of $v_i$.

\subsection{Sentence Selection via Submodularity}

The selected subset $A$ should be representative and should cover other unselected sentences in the whole set $V$. We associate a nonnegative cost $c(s)$ to each sentence $s$. We introduce a weighted coverage term for selecting sentences:

\setlength\abovedisplayskip{1pt}
\setlength\belowdisplayskip{1pt}
\begin{align}
\mathcal{H}(A) = \sum_{i\in V}\max_{j\in A \subseteq V}w_{i,j} \quad s.t. \quad c(A) \leq B \label{eq1}
\end{align}
where $c(A)=\sum_{s\in A}c(s)$ denotes the total cost of selecting $A$, and $B$ is a budget for selecting sentences. Maximizing this term encourages the selected subset $A$ to be representative and compact.

In addition, the selected sentences should be diverse. We used the diversity term introduced in~\cite{Lin:2011}: $\mathcal{Q}(A)=\sum^K_{k=1}\sqrt{\sum_{j\in \cap P_k}\frac{1}{|V|}\sum_{i\in V}w_{i,j}}$, where $\{P_1...P_K\}$ is a partition of $V$ and $|V|$ is the number of elements in $V$.

We combine two terms to obtain the final objective function: $\max_{A}\mathcal{L}(A) = \max_A\mathcal{H}(A) + \lambda\mathcal{Q}(A) \quad s.t. \quad A \subseteq V, c(A) \leq B$. The objective function is submodular and monotonically increasing. We solve the problem via a greedy algorithm. Given the selected sentences $A_{i-1}$ from step $i-1$ during optimization, in step $i$, we select the element $s_i$ with the highest marginal gain: $s_i = \argmax_{s\in V\setminus A_{i-1}}\frac{\mathcal{L}(A_{i-1}\cup s) - \mathcal{L}(A_{i-1})}{c(s)}$. The marginal gain takes the element cost $c(s)$ into account. The element cost $c(s)$ of sentence $s$ is related to its position $m$ in a document. It is defined as: $c(s)=\frac{|V|}{|V|-m+1}$.

The greedy algorithm is guaranteed to find a solution that is at least ($1-1/e$) of the optimal solution as proved in~\cite{Nemhauser:1978}, but with a complexity of $O(|V|^2)$. The optimization steps can be further accelerated using a lazy greedy approach~\cite{Leskovec:2007}. We construct a max-heap for all elements in $V$, then evaluate them in the max-heap order. With this approach, the time complexity becomes $O(|V|\log|V|)$ instead of quadratic.

\vspace{0.1cm}
\subsection{Text Semantic Similarity}
\vspace{0.1cm}

The edge weight in $G$ serves as the similarity between sentences. We compute the similarity between two sentences $i$ and $j$ by:
\begin{align}
r_{i,j}=\frac{\sum_{w \in s_i}\mathcal{S}(w, s_j) }{2|s_i|} + \frac{\sum_{w \in s_j}\mathcal{S}(w,s_i)}{2|s_j|} \label{eq2}
\end{align}
where $\mathcal{S}(w, s_j)$ is the maximal cosine similarity between input word $w$ and any words in the sentence $s_j$. Function words are filtered out when computing the similarity. This similarity value $r_{i,j}$ measures the semantic overlap between two sentences. Then we compute the distance between these two sentences for the similarity graph via: $d(x_i, x_j) = 1-r_{i,j}$.

\vspace{0.1cm}
\subsection{Combination of Different Features}
\vspace{0.1cm}
In order to leverage the strengths of deep word embeddings and n-grams features, we combine them by: (1)~\textbf{Graph fusion}: The weight assigned to each edge in a similarity graph is computed by the similarity measure between pairwise sentences. We combine the graphs from different features by using a simple weighted average of edge weights. (2)~\textbf{Late fusion}: The ranking lists from different features are combined by the popular Borda count algorithm~\cite{Dwork:2001}.

\vspace{0.1cm}
\subsection{Sentence Compression}
\vspace{0.1cm}

In order to obtain compressed or summarized form of sentences, which could then be fed into our unsupervised extractive algorithm, we trained a standard Transformer model (both encoder and decoder were composed of six stacked layers). Transformer is a neural seq2seq architecture that has shown promising results for many seq2seq tasks. We applied it to the problem of sentence compression. We also used byte pair encoding for subword segmentation~\cite{Sennrich:2015}, in order to handle unseen words (and named entities etc.) at the time of decoding.

\section{Experiments}

Our approach is evaluated on a multi-document summarization dataset: DUC-04 and two single-document datasets: CNN/DM news~\cite{nallapati2016abstractive} and NYT50~\cite{durrett2016learning}.


\vspace{0.1cm}
\subsection{Multi-Document Summarization}
\vspace{0.1cm}

The DUC-04 dataset was constructed for the multi-document summarization task using English news articles with multiple reference summaries. There are $50$ document clusters with 10 documents per cluster. For the evaluation, we used ROUGE-1 F-score (\textbf{F-1}) and Recall (\textbf{R})\footnote{ROUGE-1.5.5 with options -a -c 95 -b 665 -m -n 4 -w 1.2}. The summary length is $665$ bytes per summary.

\textbf{Baselines} We compare our approach with eight baselines. \textbf{LEAD}~\cite{rossiello:2017} simply uses the first $665$ bytes from the most recent document in each cluster. {Peer65} is the winning system in DUC-04. \textbf{Centroid}~\cite{rossiello:2017} uses word-embeddings for summarization. Three unsupervised summarization methods are also compared: \textbf{Submodular}~\cite{Lin:2011}, \textbf{MCKP}~\cite{Takamura:2009} and \textbf{LexRank}~\cite{erkan2004lexrank}. Another two methods that learn sentence embeddings are compared. \cite{cao2015ranking} uses recursive neural networks (\textbf{RNN}) and \cite{cao2015learning} uses convolutional neural networks (\textbf{CNN}) for learning sentence embeddings.

We include the results of our approach using different similarity measures with word embeddings: (1) \textbf{BERT}: the sentence embedding is computed by using the mean of word embeddings from the pretrained BERT model. The pairwise similarity between sentences is the cosine similarity. (2)\textbf{W2V}: Similar to BERT, the embeddings from word2vec model are used. Note that we did not fine-tune BERT or W2V embeddings. (3) \textbf{W2V-WMD}: the sentence similarity measure is the word mover distance introduced in~\cite{Kusner:2015}. (4) \textbf{W2V-TSS}: The text semantic similarity measure in equation~(\ref{eq2}) is used. (5) \textbf{GraphFusion}/\textbf{LateFusion}: tf-idf, BERT, W2V-WMD and W2V-TSS are combined.

We summarize the results that use different features and compare our results with those from state-of-the-art approaches in Table~\ref{tb1}. The CNN/RNN models achieve better results than our BERT and W2V models. This is because they are trained on the DUC2001 and DUC2002 datasets, while our approach is totally unsupervised and uses off-the-shelf neural word BERT or W2V embeddings only without any fine-tuning. Our results using graph fusion are better than the results of other approaches including \cite{erkan2004lexrank}, \cite{cao2015ranking}, \cite{cao2015learning} and comparable to \cite{Lin:2011}.

\begin{table}
\centering
\begin{tabular} {|c|c|c|}
\hline
Methods & F-1 & R  \\
\hline
LEAD & - & 32.4 \\
Peer65 & - & 38.2 \\
Centroid~\cite{rossiello:2017} & - & 38.8 \\
Submodular~\cite{Lin:2011} & 38.9 & 39.3  \\
MCKP~\cite{Takamura:2009} & - & 38.5 \\
LexRank~\cite{erkan2004lexrank} & - & 37.9 \\
\hline
RNN~\cite{cao2015ranking} & - & 38.8 \\
CNN~\cite{cao2015learning} & - & 38.9 \\
\hline
Ours (tf-idf) & 37.7 & 38.2 \\
Ours (W2V) & 36.9 & 37.2 \\
Ours (W2V-WMD) & 37.7 & 38.0 \\
Ours (W2V-TSS) & 37.7 & 38.1 \\
Ours (BERT) & 37.8 & 38.2 \\
Ours (LateFusion) & 37.8 & 38.2 \\
Ours (GraphFusion) & 38.8 & 39.3 \\
\hline
Ours (W2V-TSS) with compression & 38.1 & 38.7 \\
Ours (BERT) with compression & 37.9 & 38.4 \\
Ours (GraphFusion) with compression & \textbf{39.0} & \textbf{39.6} \\
\hline
\end{tabular}
\vspace{0.2cm}
\caption{Document summarization performance on the DUC2004 dataset.}
\label{tb1}
\end{table}

\begin{table}
\centering
\begin{tabular} {|c|c|c|c|}
\hline
Methods & R-1 & R-2 & R-L  \\
\hline
~\cite{Rush:2015} & 29.78 & 11.89 & 26.97 \\
~\cite{Chopra:2016} & 33.78 & 15.97 & 31.15 \\
~\cite{nallapati2016abstractive}	& 35.30	& 16.64	& 32.62 \\
Ours	& \textbf{37.12} & \textbf{18.66} & \textbf{34.38} \\
\hline
\end{tabular}
\vspace{0.2cm}
\caption{Sentence compression performance on Gigaword dataset.}
\label{tb2}
\end{table}

\begin{table}
\centering
\begin{tabular} {|c|c|c|c|}
\hline
Methods & R-1 & R-2 & R-L  \\
\hline
~\cite{Rush:2015} & 28.18 & 8.49 & 23.81 \\
~\cite{Chopra:2016} & 28.97 & 8.26 & 24.06 \\
~\cite{nallapati2016abstractive} & 28.61 & 9.42	& 25.24   \\
Ours (Gig only) & 29.04 & 10.04 & 25.74 \\
Ours (Gig+Goog)	& \textbf{29.59} & \textbf{10.89} & \textbf{26.34} \\
\hline
\end{tabular}
\vspace{0.2cm}
\caption{Sentence compression performance on DUC-2004 dataset.}
\label{tb3}
\end{table}

\vspace{0.1cm}
\subsubsection{Sentence Compression}
\vspace{0.1cm}

We used Gigaword sentence compression dataset~\cite{Rush:2015} to train the Transformer model. Gigaword dataset comprises nearly 3.9M training sentence pairs (first lines of Gigaword news articles paired with the headlines). We also used byte pair encoding for subword segmentation. In order to determine the efficacy of trained model, we used the $1951$ sentence pairs from Gigaword test set, as well as the $500$ sentence pairs from DUC-2004 sentence compression dataset~\cite{Rush:2015}. Our results on Gigaword beat the current sentence compression baselines by nearly 2 points absolute on F-scores of ROUGE-1, ROUGE-2 and ROUGE-L metrics (Table-~\ref{tb2}). On DUC-2004, we get additional improvements on the three variants of ROUGE metrics by using publicly released subset of Google sentence compression dataset~\cite{Filippova:2015} in addition to Gigaword dataset. Google compression dataset comprises nearly 200K sentence pairs (we used 180K pairs as train set, and 20K as validation set). In all metrics, on DUC-2004 dataset, we get 1 point absolute improvement on the three ROUGE metrics over current baselines (Table-~\ref{tb3}). 

Our approach to summarization uses the compressed sentences from a document to do sentence selection for document-level summarization. With sentence compression, the document summarization performance of our approach is further improved and outperforms other compared approaches as shown in Table~\ref{tb1}. The sentence compression model used to aid in document-level summarization used only the Gigaword dataset for training. We did not see any additional improvements on DUC-2004 dataset by using the additional Google compression dataset.

\begin{table*}
\centering
\begin{tabular} {|c|c|c|c|c|c|c|}
\hline
\multirow{2}{*}{Methods} & \multicolumn{3}{|c|}{CNN/DM} & \multicolumn{3}{|c|}{NYT}  \\
 & R-1 & R-2 & R-L  & R-1 & R-2 & R-L \\
\hline
ORACLE & 54.7 & 30.4 & 50.8 & 61.9 & 41.7 & 58.3 \\
LEAD3  & 40.3 & 17.7 & 36.6 & 35.5 &  17.2 & 32.0  \\
Pointer~\cite{see2017get} & 39.5 & 17.3 & 36.4 & 42.7 & 22.1 & 38.0 \\
Refresh~\cite{narayan2018ranking} & 41.3 & 18.4 & 37.5  & 41.3 &  22.0 & 37.8 \\
\hline
Ours (tf-idf) & 38.8 & 16.9 & 31.8  & 37.6 & 17.9 & 30.8 \\
Ours (W2V) & 37.4 & 16.0 & 30.6 & 36.8 & 17.1 & 29.7 \\
Ours (W2V-WMD) & 39.0 & 16.6 & 31.9 & 37.5 & 17.5 & 30.1 \\
Ours (W2V-TSS) & 38.7 & 16.7 & 31.7 & 37.7 & 17.7 & 30.2 \\
Ours (BERT) & 38.9 & 16.8 & 31.6 & 38.4 & 18.3 & 31.1 \\
Ours (GraphFusion) & 39.0 & 16.8 & 32.0 & 38.9 & 18.8 & 31.7 \\
Ours (LateFusion) & 39.2 & 17.1 & 32.2 & 39.0 & 18.8 & 31.5 \\
\hline
\end{tabular}
\vspace{0.2cm}
\caption{Document summarization performance on the CNN/DM dataset.}
\label{tb4}
\end{table*}

\subsection{Single-Document Summarization}

The CNN/DM dataset consists of online news articles from CNN and Daily Mail websites. The corpus contains a total of 287,226 article-summary pairs out of which 13,368 pairs are used for validation, 11,490 articles as test pairs and the remaining for training. However, we use about 13 thousand validation pairs for tuning our meta parameters and completely ignored the training set. The NYT50 dataset is a subset of the New York Times corpus introduced by \cite{durrett2016learning}. We use a subset of the documents that have summaries with at least 50 words, a subset known as NYT50. The final test dataset includes 3,452 test examples out of the original 9,706 articles. We evaluate these two datasets in terms of ROUGE-1 (\textbf{R-1}), ROUGE-2 (\textbf{R-2}) and ROUGE-L (\textbf{R-L}) F-scores\footnote{ROUGE-1.5.5 with options -a -c 95 -m -n 4 -w 1.2}. For both datasets, we use a budget of three sentences per summary\footnote{Please note that we don't report the summarization results with sentence compression on the CNN/DM dataset, since a compressed sentence may lose some information, and the final performance may not be improved with the constraint on the number of selected sentences.}.

\textbf{Baselines} We compare our approach with two state-of-the-art supervised learning methods: Pointer~\cite{see2017get} and Refresh~\cite{narayan2018ranking}, We also provide results from the extractive oracle system which maximizes the ROUGE score against the reference summary, and the LEAD-3 baseline that creates a summary by selecting the first three sentences in a document.

The results on both datasets are summarized in Table~\ref{tb4}. On both datasets, the results of deep features are marginally better than those of the tf-idf features. Note that our approach is unsupervised and does not use the training data, our results are surprisingly comparable to the results from the supervised learning methods including~\cite{see2017get} and~\cite{narayan2018ranking}.

\section{Conclusions}

We explore two popular deep word embeddings for the extractive document summarization task. Compared with tf-idf based features, deep embedding features are better in capturing the semantic similarity between sentences and achieve better document summarization performance. The sentence similarity measure is further improved by combining the word embeddings with n-gram features. A Transformer based sentence compression model is introduced and evaluated with our summarization approach, showing improvement in summarization performance on the DUC04 dataset. Our summarization approach is unsupervised but achieves comparable results to the supervised learning methods on the CNN/DM and NYT datasets.

\section*{Acknowledgement}
This work was supported by the Intelligence Advanced Research Projects Activity (IARPA) via Department of Defense US Air Force Research Laboratory contract number FA8650-17-C-9118.

{\small
\bibliographystyle{ieee}
\bibliography{lrec-UDS}
}

\end{document}